\documentclass[runningheads]{llncs}
\usepackage{graphicx}

\usepackage{tikz}
\usepackage{comment}
\usepackage{amsmath,amssymb} 
\usepackage{color}

\usepackage{multirow}
\usepackage{subfigure}
\usepackage{color}
\usepackage{pifont}
\usepackage{float}
\usepackage{caption}
\usepackage{bm}
\PassOptionsToPackage{hyphens}{url}\usepackage{hyperref}

\begin{document}
\pagestyle{headings}
\mainmatter
\def\ECCVSubNumber{471}  

\title{Component Divide-and-Conquer for Real-World Image Super-Resolution}

\titlerunning{Component Divide-and-Conquer for Real-World Image Super-Resolution}
%
\author{Pengxu Wei\inst{1} \and
Ziwei Xie\inst{1} \and
Hannan Lu\inst{2} \and
Zongyuan Zhan\inst{1} \and \\
Qixiang Ye\inst{3} \and
Wangmeng Zuo\inst{2} \and
Liang Lin\thanks{Corresponding Author}\inst{1,4}
}
\authorrunning{Pengxu Wei et al.}
%
\institute{Sun Yat-sen University, Guangzhou, China\\
 \and
Harbin Institute of Technology, Harbin, China\\
 \and
University of Chinese Academy of Sciences, Beijing, China\\
\and
DarkMatter AI\\
\email{weipx3@mail.sysu.edu.cn}
\email{xiezw5@mail2.sysu.edu.cn}
\email{hannanlu@hit.edu.cn}
\email{zhanzy178@gmail.com}
\email{qxye@ucas.ac.cn}
\email{wmzuo@hit.edu.cn}
\email{linliang@ieee.org}
}
%
\maketitle

\begin{abstract}
In this paper, we present a large-scale Diverse Real-world image Super-Resolution dataset, \emph{i.e.}, DRealSR, as well as a divide-and-conquer Super-Resolution (SR) network, exploring the utility of guiding SR model with low-level image components.
DRealSR establishes a new SR benchmark with diverse real-world degradation processes, mitigating the limitations of conventional simulated image degradation.
In general, the targets of SR vary with image regions with different low-level image components, \emph{e.g.}, smoothness preserving for flat regions, sharpening for edges, and detail enhancing for textures. Learning an SR model with conventional pixel-wise loss usually is easily dominated by flat regions and edges, and fails to infer realistic details of complex textures.
We propose a Component Divide-and-Conquer (CDC) model and a Gradient-Weighted (GW) loss for SR.
Our CDC parses an image with three components, employs three Component-Attentive Blocks (CABs) to learn attentive masks and intermediate SR predictions with an intermediate supervision learning strategy, and trains an SR model following a divide-and-conquer learning principle. Our GW loss also provides a feasible way to balance the difficulties of image components for SR.
Extensive experiments validate the superior performance of our CDC and the challenging aspects of our DRealSR dataset related to diverse real-world scenarios. Our dataset and codes are publicly available at {\url{https://github.com/xiezw5/Component-Divide-and-Conquer-for-Real-World-Image-Super-Resolution}}

\end{abstract}

\begin{figure*}[t]
  \centering
  \includegraphics[width=1\linewidth]{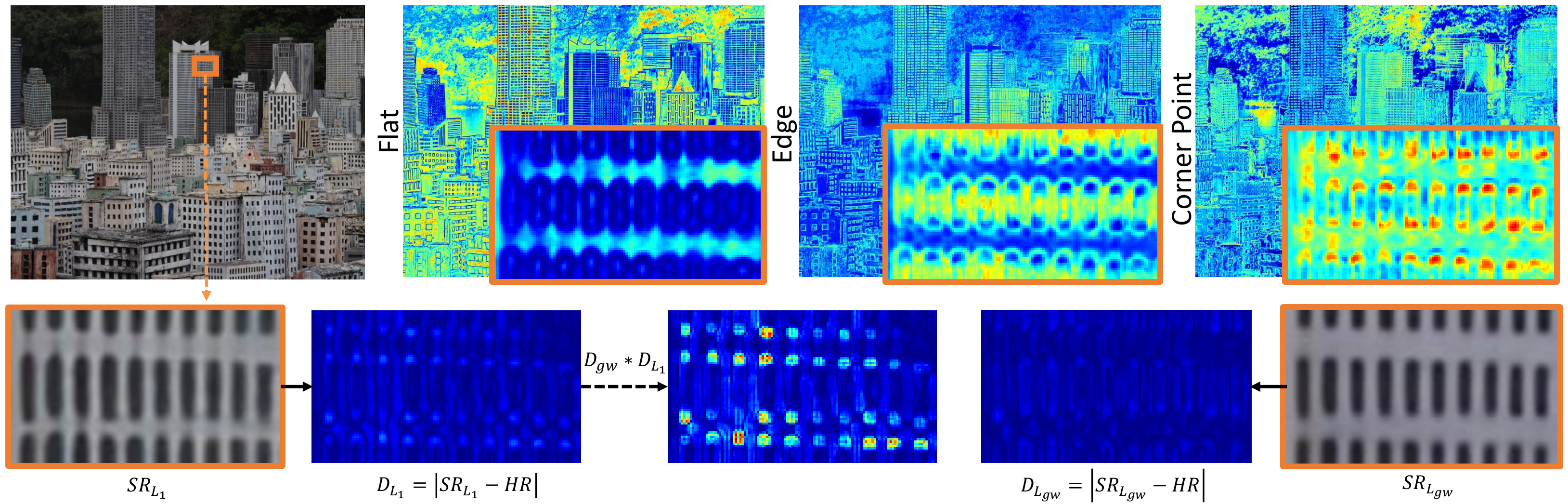}
  \caption{Image degradation reflected by loss values. $SR_{L_1}$ and $SR_{L_{gw}}$ are SR results trained with $L_1$ loss and our GW loss, respectively. The difference map $D_{L_1}$ which presents different reconstruction difficulties demonstrates the complex image degradation. It is observed that regions from small to large values in $D_{L_1}$ are relatively consistent with flat, edges and corner regions, which motivates us to explore these components and introduce a weighting strategy for $D_{L_1}$ which drives models to attend to hard regions. The first row shows three attentive masks learnt by our CDC which well predict the confidence of flat, edges and corner regions, respectively.
  }
  \vspace{-4mm}
  \label{fig:regions}
\end{figure*}

\begin{figure}[h]
  \centering
  \includegraphics[width=1\linewidth]{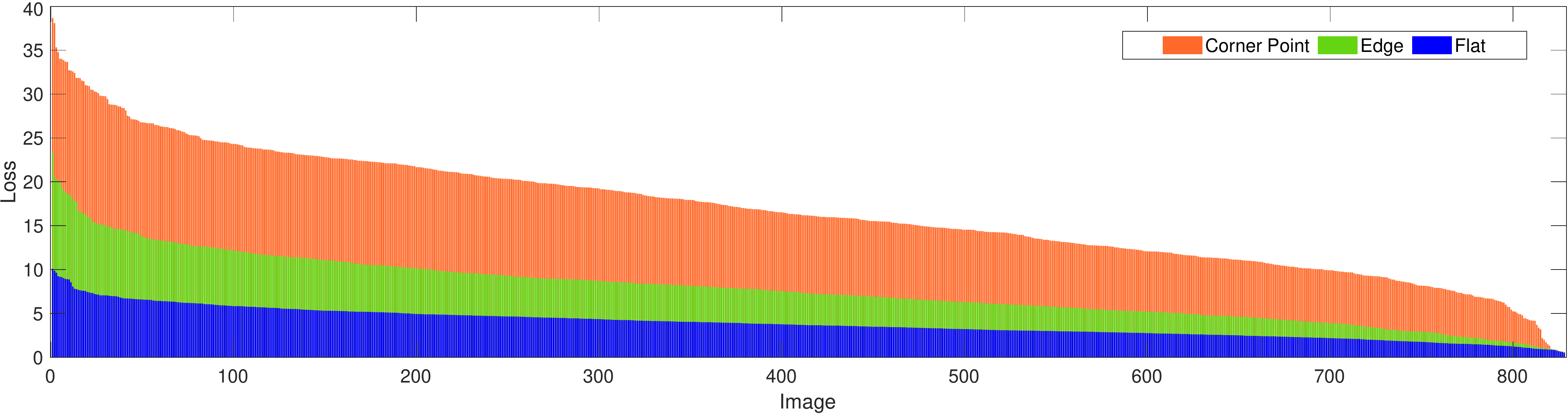}
  \caption{Component analysis for real SR. To investigate the challenging aspects, we analyze proportions of three components (flat, edges and corners) for $L_1$ loss in EDSR~\cite{EDSR} and evaluate their respective effects for SR reconstruction with averaged pixel-wise loss. Three components are observed to have different recovery difficulties: smooth regions and edges have a lower loss while corner points have a higher loss.
}
  \label{fig:cp_loss}
\end{figure}

\section{Introduction}
Single image Super-Resolution (SR) is an inherently ill-posed inverse problem that reconstructs High-Resolution (HR) images from Low-Resolution (LR) counterparts with image quality degradations.
As a fundamental research topic, it has attracted a long-standing and considerable attention in the computer vision community~\cite{glasner2009super}\cite{yang2010image}.
SR methods based on Convolutional Neural Network (CNN) (\emph{e.g.}, SRCNN \cite{SRCNN}, SRGAN \cite{SRRESNET}, EDSR \cite{EDSR}, ESRGAN \cite{ESRGAN} and RCAN \cite{RCAN}) have achieved a remarkable improvement over conventional SR methods~\cite{glasner2009super}.

However, such improvements remain limited for real-world SR applications.
The first reason is that SR models have to be trained on datasets with simulated image degradation, as LR images are obtained by simplified downsampling methods (\emph{e.g.}, bicubic) due to the difficulty of HR-LR pair collection. 
Such simulated degradation usually deviates from real ones, making the learned model not applicable to many real-world applications~\cite{zoomlearn,cai2019toward}.
The second reason is that homogenous pixel-wise losses (\emph{e.g.}, MSE) would lead to model overfitting or attend to regions for easy reconstruction.
Intuitively, the targets of SR vary with LR regions with different low-level image elements, \emph{e.g.}, smoothness preserving for flat regions, sharpening for edges, and detail enhancing for textures.
Considering that flat regions and edges are the most frequent in an image,
the models learned by homogeneous pixel-wise loss prefer addressing flat regions and edges, but usually fail to infer realistic details of complex textures. In Fig.~\ref{fig:regions}, an SR image from EDSR~\cite{EDSR} trained with $L_1$ loss presents different reconstruction difficulties in different regions, specifically in flat, edge and corner point regions. In Fig.~\ref{fig:cp_loss}, we analyze proportions of three components (flat, edges and corners) for $L_1$ loss in EDSR~\cite{EDSR} and evaluate their respective effects for SR reconstruction with averaged pixel-wise loss. Three components are observed to have different recovery difficulties: smooth regions and edges have lower loss while corner points have higher loss. Thus, these observations inspire us to investigate the utility of these three components in the SR task.

In this paper, we establish a large-scale Diverse Real-world SR benchmark, DRealSR, and propose a Component Divide-and-Conquer model (CDC) to address real-world SR challenges, \emph{i.e.}, (i) the gap between simulated and real degradation processes and (ii) the diverse image degradation processes caused by different devices.
CDC is inspired by the mechanism of Harris corner point detection algorithm~\cite{rosten2008faster}.
An image can be disentangled into three low-level components (\emph{i.e.}, flat, edge and corner) with respect to the importance of the information they convey.
Flat regions have almost constant pixel values, edges can be regarded as the boundary of different flat regions, and multiple edges interweave into corners.
In CDC, three low-level elements which facilitate an implicit composition optimization are treated as guidance to regularize the SR task.

Specifically, we first develop a base model, named HGSR, based on a stacked hourglass network. HGSR learns multi-scale features with repeated bottom-up and top-down inference across all scales. 
With HGSR, CDC builds three Component-Attentive Blocks (CABs) which are associated with flat, edges and corners, respectively. Each CAB focuses on learning one of the three low-level components with the Intermediate Supervision (IS) strategy. CDC takes the flat regions, edges and corners extracted from HR images only in the training stage and then incorporates them separately into three different branches with CABs. These three CABs form a progressive paradigm and are aggregated to yield the final SR reconstruction. 
Considering that different image regions convey different gradients in all directions, we propose a Gradient-Weighted (GW) loss function for SR reconstruction.
More complex a region is, larger impacts on the loss function it has.
Our GW loss, in a way like Focal loss~\cite{lin2017focal} for training object detectors, adapts the model training based on different image reconstruction difficulties.

In brief, our contributions are summarized as follows:
\begin{itemize}
\item
A large-scale real-world SR benchmark (DRealSR), which is collected from five DSLR cameras. DRealSR mitigates the limits of conventional simulated image degradation and establishes a new SR benchmark related to real-world challenges.

\item
A Component Divide-and-Conquer model (CDC), which, inspired by corner point detection, aims at addressing real-world SR challenges in a divide-and-conquer manner. CDC employs three Component-Attentive Blocks to learn attentive masks for different components and predicts intermediate SRs with an intermediate supervision learning strategy.

\item
A Gradient-Weighted loss, which fully utilizes image structural information for fine-detailed image SR. GW loss explores the imbalance learning issue for different image regions in the pixel-wise SR task and provides a promising solution, which can be extended to other low-level vision tasks.
\end{itemize}

\section{Related Work}
\noindent
\textbf{Datasets.}
In the area of SR, widely-used SR datasets include Set5 \cite{set5}, Set14 \cite{set14}, BSD300 \cite{BSD300}, Urban100 \cite{Urban100} and DIV2K \cite{div2k}.
Due to the difficulty of collecting HR-LR pairs, non-blind SISR approaches usually adopt a simulated image degradation for training and testing, \emph{e.g.}, bicubic downsampling.
Consequently, images in those SR datasets are usually regarded as HR images whose LR counterparts are obtained by HR downsampling.
However, the real-world degradation process can be much more complex and even nonlinear.
%
%
This simulated degradation limits related SR researches to a rather ideal SR simulation with approximately linear kernels and causes a great gap for practical SR applications \cite{cai2019toward}\cite{city100}\cite{zhang2018learning}\cite{zoomlearn}.

To fill this gap, City100 dataset \cite{city100} with 100 aligned image pairs is built for SR modeling in the realistic imaging system. However, City100 is captured for the printed postcards under an indoor environment. To capture real-world natural scenes, SR-RAW dataset is introduced for super-resolution from raw data via optical zoom \cite{zoomlearn}. RealSR dataset \cite{cai2019toward} provides a well-prepared benchmark for real-world single image super-resolution, which is captured with two DSLR cameras. In this work, we build a larger and more challenging real SR dataset with five DSLR cameras, with the target to further explore SR degradation in real-world scenarios.

\noindent
\textbf{Methods.}
Recent years have witnessed an evolution of image super-resolution research with widely-explored deep learning, which has significantly improved SR performance against traditional methods~\cite{SRCNN}. Sequentially, deep SR networks derived from various CNN models, \emph{e.g.}, VDSR \cite{VDSR}, EDSR \cite{EDSR}, SRRestNet \cite{SRRESNET}, LapSR \cite{LAPSR} and RCAN \cite{RCAN}, are presented to further improve the SR performance.
To regularize the model for ill-posed SR problem, several works are suggested to incorporate image priors, \emph{e.g.}, edge detection \cite{dai2007bilateral}\cite{fan2011single}, texture synthesis \cite{sajjadi2017enhancenet} or semantic segmentation \cite{SFT2018}. Despite their progress, most existing approaches are still tested on synthesized SR datasets with bicubic downsampling or downsampling after Gaussian blurring, while few researches are devoted to real-world SR problems.

Recently, a contextual bilateral loss (CoBi) is introduced to mitigate the misalignment issue in a real-world SR-RAW dataset \cite{zoomlearn}. Besides, LP-KPN~\cite{cai2019toward} proposes a Laplacian pyramid based method to deal with the non-uniform blur kernels for SR.
%
However, it remains limited in considering the complexity and diversity of real degradation processes among different devices, hindering the applications of real-world SR. In this work, we neither train an SR model by treating uniformly all the pixels/regions/components in an image nor bias towards only edges or textures. We parse an image into three low-level components (flat, edge and corner), explore their different importance, develop a CDC model in a divide-and-conquer learning framework and propose a GW loss to adaptively balance the pixel-wise reconstruction difficulties.

\section{DRealSR: A Large-scale Real-world SR Dataset}
To further explore complex real-world SR degradation, we build a large-scale diverse SR benchmark, named DRealSR, by zooming DSLR cameras to collect real LR and HR images. DRealSR covers 4 scaling factors (\emph{i.e.}, $\times 1$$\sim$$\times 4$).

\noindent
\textbf{Dataset collection.}
These images are captured from five DSLR cameras (\emph{i.e.,} Canon, Sony, Nikon, Olympus and Panasonic) in natural scenes and cover indoor and outdoor scenes avoiding moving objects, \emph{e.g.}, advertising posters, plants, offices, buildings, \emph{etc}.
For each scaling factor, we adopt SIFT method \cite{lowe2004distinctive} for image registration to crop an LR image to match the content of its HR counterpart. To refine the registration results, an image is cropped into patches and an iterative registration algorithm and brightness match are employed. To better facilitate the model training, considering that their image sizes are 4,000$\times$6,000 or 3,888$\times$5,184, these training images are cropped into 380$\times$380, 272$\times$272, 192$\times$192 patches for $\times 2$$\sim$$\times 4$, respectively. Since the misalignment between HR and LR possibly induces to severely blurry SR results, after each step of registration, we conduct a careful manual selection for patches. 
More details on the dataset construction are provided in the supplementary file.

\noindent
\textbf{Challenges in real-world SR.}
Due to the difficulty to capture high-resolution and low-resolution image pairs in real world, extensive SR methods are demonstrated on datasets with simulated image degradation (\emph{e.g.}, bicubic downsampling).
Compared with simulated image degradation, real-world SR exhibits the following new challenges.

\begin{itemize}
\item \textbf{More complex degradation against bicubic downsampling}. Bicubic downsampling simply applies the bicubic downsampler to an HR image to obtain the LR image. In real scenarios, however, downsampling usually is performed after anisotropic blurring, and signal-dependent noise may also be added. Thus, the acquisition of LR images suffers from both blurring, downsampling and noise. Also it is affected by in-camera signal processing (ISP) pipeline. This non-uniform property of realistic image degradation can be verified based on the reconstruction difficulty analysis of different image regions/components, Fig.\ref{fig:cp_loss}. Usually, SR models trained on bicubic degradation exhibit poor performance when handling real-world degradation.

\item \textbf{Diverse degradation processes among devices.}
In practical scenarios, differences among lens and sensors of cameras determine the different imaging patterns, which is the primary reason for explaining the diverse degradation processes in real-world SR.
Consequently, SR models learned on a real LR dataset may generalize poorly to other datasets and real-world LR images, raising another challenging issue for applications SR.
\end{itemize}

Nonetheless, different kinds of regions exhibit robustness characteristics to degradation.
For example, a flat region is less affected by the diversity of the degradation process while changes in degradation settings produce quite different results for regions with edges and corners, Fig.~\ref{fig:cp_loss}. Accordingly, this motivates us to parse an image into flat, edges and corners for easing model training. 


\begin{figure*}[t]
  \centering
  \includegraphics[width=1\linewidth]{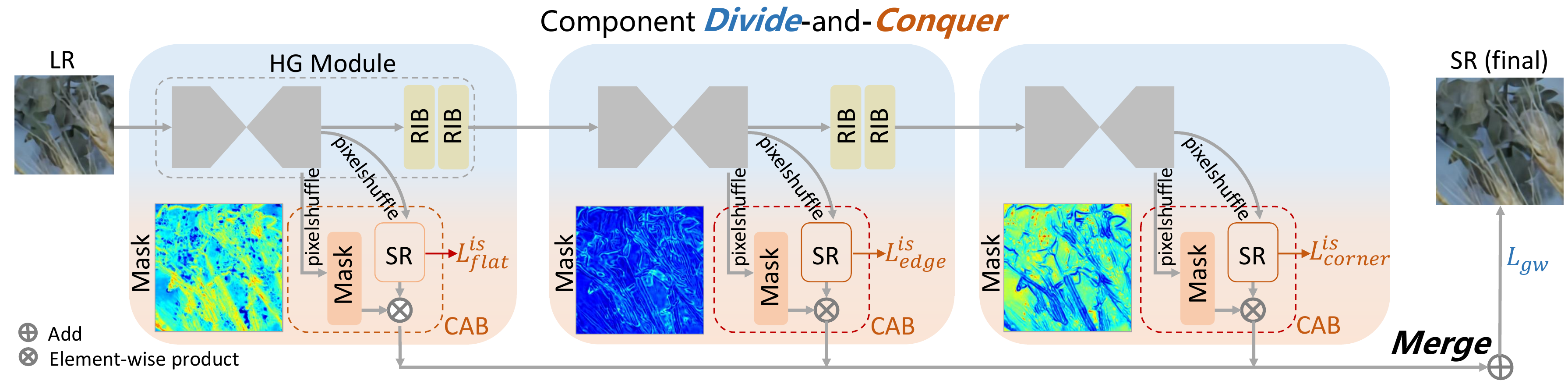}
  \caption{CDC framework. The stacked architecture enables CDC to incorporate flat regions, edges and corners in three CABs separately. Each CAB branch produces an attentive mask and an intermediate SR. This mask regularizes the produced intermediate SR to collaborate with other branches and seamlessly blend them to yield natural SR images.}
  \label{fig:framework}
\end{figure*}

\section{Real-word Image Super-Resolution}
To handle diverse image degradation, one intuitive solution is to learn anisotropic blur kernels. Due to complex contents in a natural image, however, it is hard to propose a universal solution to estimate anisotropic kernels. 
Inspired by Harris corner points \cite{harris1988combined}, image contents are divided into three primary visual components: flat, edges and corner points according to their gradient changes. These components are considered in our work. Because they represent the complexity of image contents, which indicates their reconstruction difficulty, as demonstrated in Fig.~\ref{fig:regions} and Fig.~\ref{fig:cp_loss}. For example, corners possess crucial orientation cues that                                                                                                                                                                                             control the shape or appearance of edges or textures ~\cite{harris1988combined} and are potentially beneficial for image reconstruction. Thus, these three components, \emph{i.e.}, flat, edges and corners, are explored to facilitate SR model training being free from the limits to diverse degradation processes.

Specifically, in this work, considering the reconstruction difficulty of different components,
we build an HGSR network with a stacked architecture and propose a Component Divide-and-Conquer model with a Gradient-Weighted loss to address the real SR problem. \emph{Divide} arranges the introduction order as flat, edges and corners to facilitate the feature learning in the network from easy to hard; \emph{conquer} separately produces intermediate SR results for each component which are \emph{merged} into the final SR prediction.
\begin{itemize}
\item \emph{Divide}: Consider the complexity of image contents in flat, edges and corner point regions, we guide three HG modules to emphatically learn component-attentive masks from LR images respectively, with the component parsing guidance from HR images.
It is noted that, we do not directly detect three components from LR images with off-the-shelf methods but predict their maps coherent with the HR image. The main reason is that the low quality of LR images hinders the more accurate corner point detection and yields undesirable detection results. Another reason is that this strategy avoids corner point detection for each image in the test stage.

\item \emph{Conquer}: Three Component-Attentive Blocks produce different component-attentive masks and intermediate SR predictions. The generated attentive maps present remarkable characteristics of three components. Meanwhile, intermediate SR results are consistent with the characteristics of three regions.

\item \emph{Merge}: 
    To yield the final SR result, we collaboratively aggregate three intermediate SR outputs weighted by the corresponding component-attentive maps. In particular, a GW loss is proposed to drive the model to adapt learning objectives to their reconstruction difficulties. 
\end{itemize}

\subsection{Formulation} 
In the real SR, given $N$ LR-HR pairs, we estimate an SR image $ {\mathbf {\hat x}_i}$ by minimizing the loss function $\mathcal L = \frac{1}{{N}}\sum\nolimits_{i = 1}^N {{\mathcal L_{rec}({\mathbf {\hat x}_i}, {\mathbf x_i})}}$, where $\mathcal L_{rec}(\cdot)$ is a reconstruction loss function. The network learns a mapping function $\mathcal F$ from the LR image $\mathbf y_i$ to the HR image $\mathbf x_i$; namely, $ {\mathbf {\hat x}_i} = \mathcal F({\mathbf y_i};\Theta )$, where $\Theta$ is the network model parameter.
In general, the realistic image degradation is complex and diverse as claimed above. To make it relatively tractable, our CDC employs three CABs and learn models with the intermediate supervision in a divide-and-conquer manner, rather than directly learning LR-HR mapping function or estimating blur kernels. Thus, our loss function is defined as follows,
\begin{equation}
\label{equ:model_loss}
\mathcal L = \frac{1}{N}\sum\nolimits_{i = 1}^N {[\mathcal L_{rec}({\mathbf {\hat x}_i},{\mathbf x_i}) + \sum\nolimits_{e = 1}^3 {{\mathcal L_{is}}(\tilde {\mathbf x}_i^e,{\mathbf x_i})} } ],
\end{equation}
where $\mathcal L_{is}$ is the intermediate loss function, the index $e$ represents an CAB module that is specific to either \emph{flat}, \emph{edge} or \emph{corner}, and $\tilde {\mathbf x}_i^e$ is the intermediate SR prediction (${{\tilde {\mathbf x}_e}}$ for simplicity in the following sections).

\subsection{Hourglass Super-Resolution Network}
We propose a basemodel, Hourglass Super-Resolution network (HGSR), which has a stacked hourglass architecture \cite{hourglass} followed by a pixelshuffle layer \cite{subpixel}. The hourglass (HG) architecture is motivated to capture information at every scale and has a superior performance for keypoint detection~\cite{hourglass}. Its hourglass module can be regarded as an encoder-decoder with skip connections to preserve spatial information at each resolution and bring them together to predict pixel-wise outputs. In the HG module, an input passes through a convolutional layer firstly and then is downsampled to a lower resolution by a maximum pooling layer. During Top-Down inference, it repeats this procedure until reaching the lowest resolution. Next, a Bottom-Up inference performs constantly upsampling by nearest neighbor interpolation and combines features across scales by skip-connection layers until the original resolution is restored.

The conventional connection between two HG modules are two Residual Blocks (RBs) \cite{hourglass}, as shown in Fig.~\ref{fig:modules}(a).
HGSR replaces RBs with Residual Inception Blocks (RIBs) \cite{szegedy2017inception}. RIBs have a parallel cascade structure and concatenate feature maps produced by filters of different sizes. Besides, HGSR utilizes the Intermediate Supervision (IS) strategy for model learning.
The main difference of the IS module in \cite{hourglass} is that HGSR does not recursively feed the IS prediction to the next HG module, as shown in Fig.~\ref{fig:modules}(b). 
The intermediate loss function $\mathcal L_{is}$ in HGSR is the $\mathcal L_1$ loss.

\subsection{Component Divide-and-Conquer Model}
Our Component Divide-and-Conquer model takes HGSR as the backbone and follows the divide-and-conquer principle to learn the model.
Specifically, CDC focuses on three image components, \emph{i.e.}, flat, edges and corners, rather than edges or/and complex textures. This makes it relatively tractable to solve the ill-posed real SR problem.
These components are explicitly extracted from HR images with Harris corner detection algorithm, separately in CABs and are implicitly blended seamlessly to yield natural SR results by minimizing a GW loss. Although the guidance of three components is from HR images, CDC infers the component probability maps in the test stage without any detection.

\begin{figure}[t]
  \centering
  \includegraphics[width=1\linewidth]{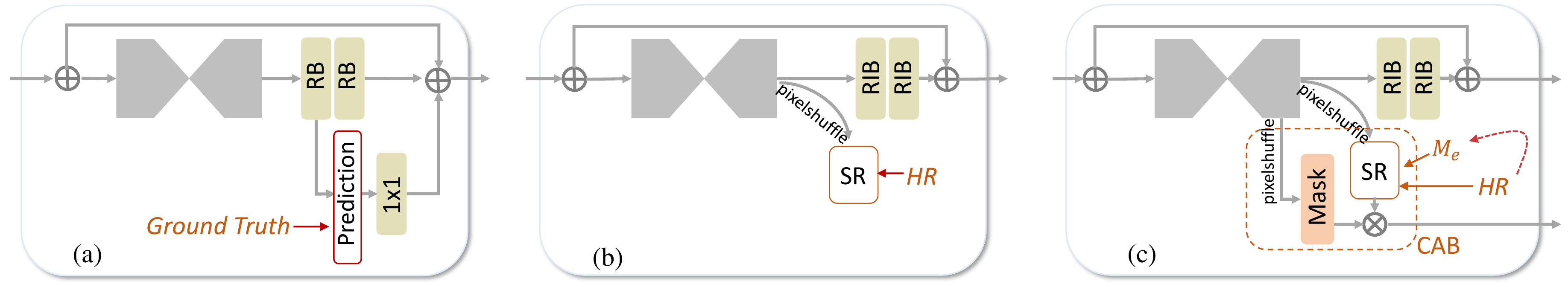}
  \caption{CAB and intermediate supervision in the HG module. We compare (a) the IS in the basic HG \cite{hourglass} with those in HGSR (b) and our CDC (c).
  In \cite{hourglass}, an intermediate prediction recursively joins into the next HG module after 1$\times$1 convolution for human pose estimation. For SR, our IS strategy in (b) and (c) avoids the recursive operation which tends to invite large disturbance for feature learning in the backbone. 
  }
  \label{fig:modules}
\end{figure}

\noindent
\textbf{Component-Attentive Block.}
CDC has three CABs which respectively correspond to either flat, edges or corners. Since it inherits the advantages of HGSR with a cascaded hourglass network, CDC is suitable to incorporate the intermediate supervision. As shown in Fig.\ref{fig:modules}(c), CAB consists of two pixel-shuffle layers. One is used to generate a coarse intermediate SR result. The other one is used to generate a mask which indicates the component probability map. It weights this coarse SR for the final SR reconstruction together with outputs of other CABs. In the training stage, CDC leverages the HR image as intermediate supervision to generate an IS loss weighted by the guidance of the component mask from HR.
Accordingly, the intermediate loss function in an CAB is defined as
\begin{equation}
\label{equ:intermediate_loss}
\mathcal L_{is} = l({{\bm M}_e}*\mathbf x,{{\bm M}_e}*{{\tilde {\mathbf x}_e}}),
\end{equation}
where $e$ is similar to that defined in Equ.~\ref{equ:model_loss} and $*$ denotes the entry-wise product;
${\bm M}_e$ is the component guidance mask extracted from HR images. In general, $l(\cdot)$ can be any loss functions; we adopt widely-used $L_1$ loss function in the CDC.

\begin{figure}[t]
  \centering
  \includegraphics[width=0.95\linewidth]{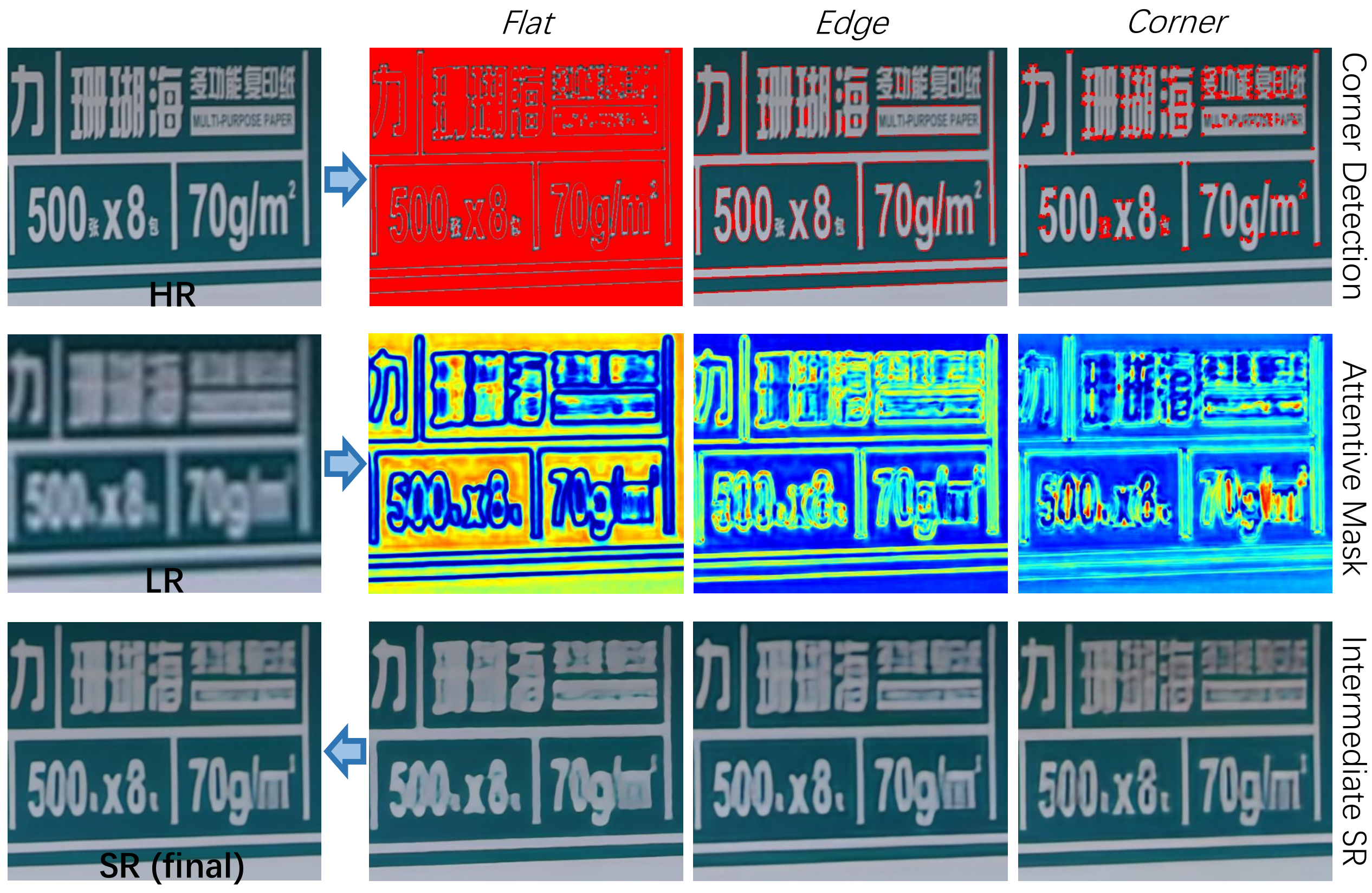}
  \caption{Harris corner point detection, learned component-attentive masks and intermediate SR images from three CABs. Component-attentive masks from each CAB present a high similarity to flat, edges and corners, respectively.}
  \label{fig:corner_mask_sr}
\end{figure}

As shown in Fig. \ref{fig:corner_mask_sr}, the learned component-attentive masks in three CABs exhibit their own characteristics in indicating flat regions, edges and textures, respectively. Accordingly, their intermediate SR results are also consistent with these characteristics. To further aggregate these three types of information, we will describe how to collaboratively aggregate them to yield the final SR result with a gradient-weighted loss.

\noindent
\textbf{Gradient-Weighted Loss.}
For conventional pixel-wise loss, regions in an image are treated identically. However, flat regions and edges dominate the loss function due to their large quantity in images.
Thus, the learned SR models incline to address flat regions and edges, but fail to infer realistic details of complex textures. Inspired by Focal loss \cite{lin2017focal}, we propose to suppress a large number of simple regions while emphasizing the hard ones.
Notably, this strategy is also crucial for low-level vision tasks.
In our work, the solution of flat, edge and corner point detection provides a plausible disentanglement of images according to their importance, which can thus be used to determine the easy and hard regions and obtain the final SR prediction $\hat {\mathbf x}$ as the sum of outputs from three CABs, namely, $\hat {\mathbf x} = \sum\nolimits_e {{\bm A_e}*{{\tilde {\mathbf x}}_e}}$, where $\bm A_e$ is a component-attentive mask.

We propose a Gradient-Weighted loss to dynamically adjust their roles for minimizing the SR reconstruction loss. Following this philosophy, the flat and single edge regions are naturally classified as simple regions. Corners are categorized as difficult regions since they possess the fine-details in images. Considering the diversity in the first-order gradient of different regions, the new reconstruction loss function for SR, named GW loss, is defined as
\begin{equation}
\label{equ:gw_loss}
\mathcal L_{gw} = l(D_{gw} * \mathbf x, D_{gw} * \hat {\mathbf x}),
\end{equation}
where $D_{gw}=(1 + \alpha {D_x})(1 + \alpha {D_y})$; ${D_x} = |G_x^{sr} - G_x^{hr}|$ and ${D_y} = |G_y^{sr} - G_y^{hr}|$ represent gradient difference maps between SR and HR in the horizontal and vertical directions; $\alpha$ is a scalar factor to determine the quantity for this weighting in the loss function. Generally, $l(\cdot)$ can be also any loss function and we adopt $L_1$ loss in this paper. If $\alpha=0$, GW loss becomes the original loss $l(\mathbf x, \hat {\mathbf x})$. $\alpha$ is 4 in our experiments. This GW loss is regarded as the reconstruction loss $\mathcal L_{rec}$.

\section{Experiments}
\subsection{Experimental settings} \label{sec:sub_exp_setting}
\noindent
\textbf{Dataset.} We conduct experiments on an existing real-world SR dataset, RealSR, and our DRealSR.
\textbf{RealSR} \cite{cai2019toward} has 595 HR-LR image pairs captured from two DSLR cameras. 15 image pairs at each scaling factor of each camera are selected randomly for building the testing set and the rest pairs are training set. Their image sizes are in the range of [700, 3100] and [600, 3500] and each training image is cropped in 192$\times$192 patches.
For $\times 2$$\sim$$\times 4$, our \textbf{DRealSR} has 884, 783 and 840 image pairs respectively, where 83, 84 and 93 image pairs are randomly selected for testing respectively and the rest are for training at each scaling factor. 

\noindent
\textbf{Network Architecture.}
CDC cascades six HG modules. In each HG module, a residual block followed by a max-pooling layer for the top-down process and nearest neighbor method for the bottom-up process. For shortcut connection across two adjacent resolutions, we also use a residual block consisting of three convolution layers: 1$\times$1, 3$\times$3 and 1$\times$1 filters. Between two hourglass modules, there are two connection layers using Residual Inception Block \cite{szegedy2017inception} for multi-scale processing. To introducing the intermediate supervision, those six HG modules are divided into three groups and the last HG in a group generates a coarse SR image by an upsampling layer of pixelshuffle.

\noindent
\textbf{Implementation Details.}
Harris Corner detection method \cite{harris1988combined} with OpenCV is used. In our experiments, we use Adam optimizer \cite{adam} and set 0.9 and 0.999 for its exponential decay rates. 
The initial learning rate is set to 2e-4 and then reduced to half every 100 epochs. For each training batch, we randomly extract 16 LR patches with the size of 48 $\times$ 48. 
All of our experiments are conducted in PyTorch. %
Three common image quality metrics are used to evaluate SR models, \emph{i.e.}, peak signal-to-noise ratio (PSNR) and structural similarity index (SSIM) \cite{ssim} and Learned Perceptual Image Patch Similarity (LPIPS) \cite{lpips}. PSNR is evaluated on the Y channel; SSIM and LPIPS are on RGB images.

\subsection{Model Ablation Study}

\noindent
\textbf{Evaluation on HG blocks.}
Our base model, HGSR, adopts a stacked hourglass network \cite{hourglass} as the backbone. We provide experimental evaluations on the number of HG blocks in Table \ref{tab:HG_study}. It is observed that the SR performance has a sustaining boost when the number of HG blocks in HGSR
increases from 2 to 4 while it has a stable performance when the number of HG blocks is larger than 6. Thus, the number of HG blocks is set 6 in our experiments.

\begin{flushleft}
\begin{minipage}[t]{\textwidth}
 \begin{minipage}[t]{0.5\textwidth}
     \makeatletter\def\@captype{table}\makeatother\caption{Evaluation of the number of HG blocks}
     \label{tab:HG_study}
     \resizebox{0.98\textwidth}{!}{
     \begin{tabular}{|l|c|ccc|}
        \hline
        \multicolumn{1}{|c|}{Method} & HG Blocks & PSNR   & SSIM  & LPIPS \\ \hline
        \hline
        \multirow{4}{*}{HGSR}        & 2         & 31.55 & 0.847 & 0.336 \\
                                     & 4         & 31.80 & 0.854 & 0.312 \\
                                     & 6         & \textbf{31.95} & \textbf{0.854} & 0.304 \\
                                     & 8         & 31.94 & 0.854 & \textbf{0.303} \\ \hline
	 \end{tabular}
     }
  \end{minipage}
  %
  \begin{minipage}[t]{0.5\textwidth}
    \makeatletter\def\@captype{table}\makeatother\caption{Ablation study of the proposed CDC}
    \label{tab:component_study}
     \resizebox{0.97\textwidth}{!}{
     \begin{tabular}{|l|ccc|}
	    \hline
        \multicolumn{1}{|c|}{Method} & PSNR   & SSIM  & LPIPS \\ \hline
        \hline
        HGSR(baseline, w/o IS)      & 31.95 & 0.854 & 0.304 \\ \hline
        HGSR(baseline)              & 32.13 & 0.855 & 0.310 \\ \hline
        HGSR+RIB                   & 32.15 & 0.857 & 0.310 \\ \hline
        HGSR+RIB+CAB         & 32.27 & 0.858 & 0.302 \\ \hline
        HGSR+RIB+CAB+GW      & \textbf{32.42} & \textbf{0.861} & \textbf{0.300} \\
        \hline
	 \end{tabular}
     }
%
  \end{minipage}
  %
  \begin{minipage}[t]{0.5\textwidth}
        \makeatletter\def\@captype{table}\makeatother\caption{Evaluation of CDC in flat, edge, and corner regions}
        \label{tab:region_study}
        \resizebox{0.98\textwidth}{!}{
        \begin{tabular}{|c|ccc|ccc|}
            \hline
            \multirow{2}{*}{Method} & \multicolumn{3}{c|}{Regions}  & \multirow{2}{*}{PSNR} & \multirow{2}{*}{SSIM} & \multirow{2}{*}{LPIPS} \\ \cline{2-4}
                                             & \textit{flat} & \textit{edge} & \textit{corner} &                        &                         &       \\  \hline
            \hline
            \multirow{7}{*}{CDC}            & \checkmark             &               &                  & 32.03        & 0.856         & 0.310      \\
                                            &                        & \checkmark    &                  & 32.25        & 0.858         & 0.307      \\
                                            &                        &               & \checkmark       & 32.37        & 0.861         & 0.302      \\
                                            & \checkmark             & \checkmark    &                  & 32.23        & 0.860         & 0.301      \\ 
                                            & \checkmark             &               & \checkmark       & 32.39        & 0.861         & 0.300      \\ 
                                            &                        & \checkmark    & \checkmark       & 32.40        & 0.861         & \textbf{0.298}      \\ 
                                            & \checkmark             & \checkmark    & \checkmark       &\textbf{32.42}   &\textbf{0.861}  & 0.300      \\  \hline
        \end{tabular}
        }
   \end{minipage}
  \begin{minipage}[t]{0.5\textwidth}
     \makeatletter\def\@captype{table}\makeatother\caption{Comparison results of our proposed GW loss with $L_1$ loss}
     \label{tab:comp_loss}
     \resizebox{1\textwidth}{!}{
     \begin{tabular}{|c|c|ccc|}
        \hline
        Method                      & Loss & PSNR  & SSIM  & LPIPS \\ \hline
        \hline
        \multirow{2}{*}{SRResNet \cite{SRRESNET}}  & $L_1$   & 31.63         & 0.847            & 0.341 \\
                                                   & $L_{gw}$    & \textbf{31.93}         & \textbf{0.853}            & \textbf{0.321} \\ \hline
        \multirow{2}{*}{EDSR \cite{EDSR}}          & $L_1$ & 32.03          & \textbf{0.855}          & 0.307          \\
                                                   & $L_{gw}$    & \textbf{32.27} & 0.857 & \textbf{0.304} \\ \hline
        \multirow{2}{*}{HGSR(Our baseline)}        & $L_1$ & 32.15          & 0.857          & \textbf{0.310}          \\
                                                   & $L_{gw}$    & \textbf{32.25} & \textbf{0.857} & 0.313 \\ \hline
        \multirow{2}{*}{CDC(Ours)}                 & $L_1$ & 32.27          & 0.858          & 0.302          \\
                                                   & $L_{gw}$    & \textbf{32.42} & \textbf{0.861} & \textbf{0.300} \\ \hline
    \end{tabular}
     }
  \end{minipage}
\end{minipage}
\end{flushleft}

\noindent
\textbf{Evaluation on IS and RIB.}
We leverage the intermediate supervision strategy to hierarchically supervise the model learning. As shown in Table \ref{tab:component_study}, HGSR with IS achieves 0.18dB PSNR gains. In the following parts, if no special claim, HGSR denotes the base model with IS. Besides, two convolution layers are added between the two HG modules in HGSR to build their connections. To aggravate multi-scale information, these two layers are substituted by two RIBs. This modification slightly improves the base model with 0.02dB PSNR gains. Besides, our CAB and the GW loss have 0.12 and 0.15 dB PSNR improvements, respectively. Particularly, our final version, \emph{i.e.}, CDC, exhibits an impressive improvement (\emph{i.e.}, 0.29dB) compared with the base model HGSR.

\noindent
\textbf{Evaluation on Component-Attentive Block.}
As demonstrated in Table \ref{tab:component_study}, our CAB brings an improvement of 0.12 dB in PSNR.
In order to analyze CAB, we conduct experiments on different guidance from flat, edge and corner regions, as shown in Table \ref{tab:region_study}. Without corner branches, our model has a significant drop of 0.19 dB by PSNR. Among three components (\emph{i.e.}, flat, edges and corners), corners that represent important information play a crucial role in the SR task, although they have a small quantity in an image. This observation is encouraging to pay more attention to exploring corner points in the SR task, as well as directly on edges or/and textures.

\noindent
\textbf{Evaluation on Gradient-Weighted Loss.}
In Table \ref{tab:comp_loss}, in comparison with $L_1$ loss, the GW loss respectively introduces 0.30 dB, 0.25 dB, 0.10 dB and 0.15 dB improvement in PSNR for EDSR, SRResNet, HGSR, and our CDC. This indicates that our proposed GW loss can be applied to other SR models to further improve their performance. Notably, the GW loss rooted in $L_1$ loss achieves a greater improvement than $L_1$. Therefore, our GW loss provides a new way to understand the SR model learning and can be explored in other loss functions and other low level vision tasks.

\begin{table*}[t]
\begin{center}
\caption{Performance comparison on RealSR~\cite{cai2019toward} and DRealSR datasets}
\label{tab:main_result}
\resizebox{1\textwidth}{!}{
\begin{tabular}{|c|c|ccc|ccc|ccc|ccc|}
\hline
\multirow{3}{*}{Method} & \multirow{3}{*}{Scale} & \multicolumn{6}{c|}{Training Set: DRealSR}                                                         & \multicolumn{6}{c|}{Training Set: RealSR}                                                            \\ \cline{3-14}
                        &                        & \multicolumn{3}{c|}{Test on RealSR \cite{cai2019toward}} & \multicolumn{3}{c|}{Test on DRealSR}           & \multicolumn{3}{c|}{Test on RealSR \cite{cai2019toward}} & \multicolumn{3}{c|}{Test on DRealSR}           \\ \cline{3-14}
                        &                        & PSNR            & SSIM           & LPIPS          & PSNR           & SSIM           & LPIPS          & PSNR            & SSIM           & LPIPS          & PSNR           & SSIM           & LPIPS          \\ \hline
Bicubic
& \multirow{7}{*}{$\times 2$} & 31.67 & 0.887 & 0.223 & 32.67 & 0.877 & 0.201 & 31.67 & 0.887 & 0.223 & 32.67 & 0.877 & 0.201   \\
SRResNet \cite{SRRESNET}
&                             & 32.65 & 0.907 & 0.169 & 33.56 & 0.900 & 0.163 & 33.17 & 0.918 & 0.158 & 32.85 & 0.890 & 0.172   \\
EDSR \cite{EDSR}
&                             & 32.71 & 0.906 & 0.172 & 34.24 & 0.908 & 0.155 & 33.88 & 0.920 & 0.145 & 32.86 & \textbf{0.891} & 0.170  \\
ESRGAN \cite{ESRGAN}
&                             & 32.25 & 0.900 & 0.185 & 33.89 & 0.906 & 0.155 & 33.80 & 0.922 & 0.146 & 32.70 & 0.889          & 0.172   \\
RCAN \cite{RCAN}
&                     & \textbf{32.88}& 0.908 & 0.173 & 34.34 & 0.908 & 0.158 & 33.83 & 0.923 & 0.147 & \textbf{32.93} & 0.889 & 0.169          \\
LP-KPN \cite{cai2019toward}
&                             & 32.14 & -     & -     & 33.88 & -     & -     & -     & -     & -     & -      & -  & -    \\
DDet \cite{shi2020ddet}
&                             & 32.58 &-      &-      & 33.92 & -     &-      & 33.22 &-      &-      & 32.77 &- &- \\
\textbf{CDC (Ours)}
&                             & 32.81 & \textbf{0.910} & \textbf{0.167} & \textbf{34.45} & \textbf{0.910} & \textbf{0.146} & \textbf{33.96}  & \textbf{0.925} & \textbf{0.142} & 32.80 & 0.888  & \textbf{0.167} \\ \hline
Bicubic
& \multirow{7}{*}{$\times 3$} & 28.63 & 0.809 & 0.388 & 31.50 & 0.835  & 0.362 & 28.61 & 0.810 & 0.389 & 31.50 & 0.835   & 0.362          \\
SRResNet \cite{SRRESNET}
&                             & 28.85 & 0.832 & 0.290 & 31.16 & 0.859  & 0.272 & 30.65 & 0.862 & 0.228 & 31.25 & 0.841   & 0.267          \\
EDSR \cite{EDSR}
&                             & 29.50 & 0.841 & 0.266 & 32.93 & 0.876  & 0.241 & 30.86 & 0.867 & 0.219 & 31.20 & 0.843   & \textbf{0.264} \\
ESRGAN \cite{ESRGAN}
&                             & 29.57 & 0.841 & 0.266 & 32.39 & 0.873  & 0.243 & 30.72 & 0.866 & 0.219 & 31.25 & 0.842   & 0.268          \\
RCAN \cite{RCAN}
&                             & \textbf{29.68}& 0.841 & 0.267 & 33.03  & 0.876 & \textbf{0.241} & 30.90& 0.864 & 0.225   & 31.76 & 0.847          & 0.268          \\
LP-KPN \cite{cai2019toward}
&                             & 29.20 & -     & -     & 32.64 & -      & -     & 30.60 & -      & -    & \textbf{31.79}& - & -          \\
DDet \cite{shi2020ddet} &     & 29.48 & -     &-      &32.13  &-       &-      & 30.62 &-       &-     & 31.77  &-       &- \\
\textbf{CDC (Ours)}               &     & 29.57 & \textbf{0.841} & \textbf{0.261} & \textbf{33.06} & \textbf{0.876} & 0.244   & \textbf{30.99}  & \textbf{0.869} & \textbf{0.215} & 31.65  & \textbf{0.847}  & 0.276    \\ \hline
Bicubic
& \multirow{7}{*}{$\times 4$} & 27.24 & 0.764 & 0.476 & 30.56 & 0.820  & 0.438 & 27.24 & 0.764          & 0.476          & 30.56          & 0.820          & 0.438          \\
SRResNet \cite{SRRESNET}&     & 27.63 & 0.785 & 0.368 & 31.63 & 0.847  & 0.341 & 28.99 & 0.825          & 0.281          & 29.98          & 0.822          & 0.347          \\
EDSR \cite{EDSR}        &     & 27.77 & 0.792 & 0.339 & 32.03 & 0.855  & 0.307 & 29.09 & 0.827          & 0.278          & 30.21          & 0.817          & \textbf{0.344} \\
ESRGAN \cite{ESRGAN}    &     & 27.82 & 0.794 & 0.340 & 31.92 & 0.857  & 0.308 & 29.15 & 0.826          & 0.279          & 30.18          & 0.821          & 0.353          \\
RCAN \cite{RCAN}        &     & 27.93 & 0.795 & 0.341 & 31.85 & 0.857  & 0.305 & 29.21 & 0.824          & 0.287          & 30.37          & 0.825          & 0.349          \\
LP-KPN \cite{cai2019toward}&  & 27.79 & -     & -     & 31.58 & -      & -     & 28.65 & -              & -              & \textbf{30.75}          & - & -          \\
DDet \cite{shi2020ddet} &     & 27.83 &-      &-      & 31.57 &-       &-      & 28.94 &-               &- &30.12 &- &- \\
\textbf{CDC (Ours)}               &     & \textbf{28.11}  & \textbf{0.800} & \textbf{0.330} & \textbf{32.42} & \textbf{0.861} & \textbf{0.300}   & \textbf{29.24}  & \textbf{0.827} & \textbf{0.278}  & 30.41  & \textbf{0.827}          & 0.357          \\ \hline
\end{tabular}
}
\end{center}
\end{table*}

\subsection{Comparison with state-of-the-arts on real SR datasets}
We compare our method with several state-of-the-art SR methods, including SRResNet \cite{SRRESNET}, EDSR \cite{EDSR}, ESRGAN \cite{ESRGAN}, RCAN \cite{RCAN}, LP-KPN \cite{cai2019toward} and DDet \cite{shi2020ddet}. Among these SR methods, LP-KPN \cite{cai2019toward} and DDet \cite{shi2020ddet} are the only two designed to solve the real-world SR problem.
Quantitative comparison results are given in Table~\ref{tab:main_result}.
LP-KPN \cite{cai2019toward} and DDet \cite{shi2020ddet} are trained on the Y channel and other methods are trained on RGB images. Considering this difference, LPIPS and SSIM of LP-KPN and DDet are not provided since these two metrics are evaluated on RGB images.
Our CDC outperforms the state-of-the-art algorithms on two real-world SR datasets. On DRealSR, CDC achieves the best results in all scales and notably improves the performance by about 0.4 dB for $\times$4. Similar to the performance on DRealSR, PSNR and SSIM of CDC on RealSR are also superior to the others, validating the effectiveness of our method.

\begin{figure*}[t]
  \centering
  \includegraphics[width=0.95\linewidth]{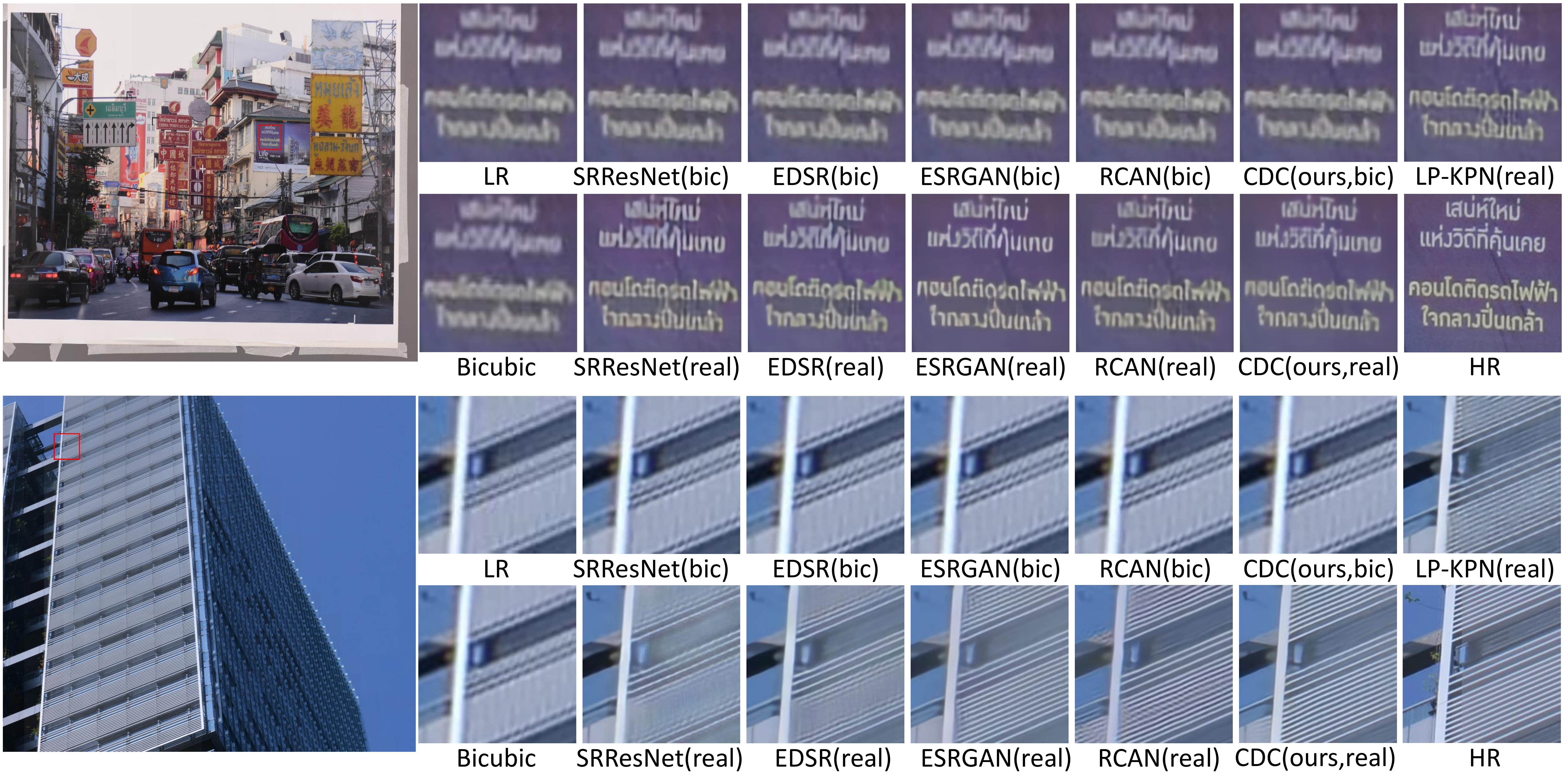}
  \caption{SR results for $\times 4$ on DRealSR in comparison with state-of-the-art approaches. `real' indicates models trained on DRealSR while `bic' indicates those trained on a dataset version by bicubic downsampling DRealSR HR images.}
  \label{fig:example_sr_ressult}
\end{figure*}

Fig.~\ref{fig:example_sr_ressult} visualizes SR results of the competing methods and ours.
It is observed that existing SR methods (\emph{e.g.}, EDSR, RCAN, LP-KPN) are prone to generate realistic detailed textures with visual aliasing and artifacts. For instance, in the second example in Fig.~\ref{fig:example_sr_ressult}, SRRestNet, EDSR and LP-KPN  produce blurry details of the building and the result of RCAN has obvious aliasing effects. In comparison, our proposed CDC reconstructs sharp and natural details.

To further validate challenges of RealSR and our DRealSR, we also conduct the cross-testing on the two datasets, \emph{i.e.}, training models on one of them and then testing on the other one. In Table \ref{tab:main_result}, CDC trained on DRealSR maintains a superior performance in all scales when tested on RealSR. However, for models trained on RealSR, the testing performance drops greatly on DRealSR, especially for $\times$4, which indicates that DRealSR is more challenging than RealSR.


\begin{table*}[t]
\small
\begin{center}
\caption{SR performance comparison upon the image degradation}
\label{tab:comp_bic_real_degradation}
\resizebox{0.8\textwidth}{!}{
\begin{tabular}{|c|c|c|ccc|ccc|}
\hline
\multirow{3}{*}{Method}                                    & \multicolumn{2}{c|}{\multirow{2}{*}{Training set}} & \multicolumn{6}{c|}{Testing set (DRealSR)}                                                        \\ \cline{4-9}
                                                           & \multicolumn{2}{c|}{}         & \multicolumn{3}{c|}{Bicubic}                     & \multicolumn{3}{c|}{Real}                        \\ \cline{2-9}
                                                           & Dataset                       & Degradation        & PSNR           & SSIM           & LPIPS          & PSNR           & SSIM           & LPIPS          \\ \hline
SRResNet \cite{SRRESNET} & DRealSR                     & Bicubic            & 41.28          & 0.954          & 0.103          & 30.61          & 0.822          & 0.422          \\
EDSR \cite{EDSR}         & DRealSR                     & Bicubic            & 41.49          & 0.956          & 0.099          & 30.60          & 0.822          & \textbf{0.421} \\
RRDB \cite{ESRGAN}       & DRealSR                     & Bicubic            & 41.66          & 0.957          & 0.097          & 30.60          & 0.822          & 0.425          \\
\textbf{CDC (Ours)}                                                 & DRealSR                     & Bicubic            & \textbf{41.78} & \textbf{0.957} & \textbf{0.096}          & \textbf{30.63} & \textbf{0.822} & 0.425          \\ \hline
SRResNet \cite{SRRESNET} & DRealSR                     & Real               & 31.43          & 0.864          & 0.249          & 31.63          & 0.847          & 0.341          \\
EDSR \cite{EDSR}         & DRealSR                     & Real               & 32.53          & 0.880          & 0.231          & 32.03          & 0.855          & 0.307          \\
RRDB \cite{ESRGAN}       & DRealSR                     & Real               & 32.37          & 0.877          & 0.234          & 31.92          & 0.857          & 0.308          \\
\textbf{CDC (Ours)}                                                 & DRealSR                     & Real               & \textbf{32.54} & \textbf{0.883} & \textbf{0.215} & \textbf{32.42} & \textbf{0.861} & \textbf{0.300} \\ \hline
\end{tabular}
}
\end{center}
\end{table*}

\subsection{Analysis on Real and Simulated SR Results}
In this section, we analyze the bicubic and real image degradation on DRealSR.
In Table~\ref{tab:comp_bic_real_degradation}, the performance of real image degradation on our dataset is very close to that of bicubic image degradation even if the training set is different. Actually, the performance on PSNR, SSIM and LPIPS is close to that of bicubic upsampling method, as shown in Table~\ref{tab:main_result}. Thus, no matter which model is used, it is not useful to restore the real image with the model trained on bicubic images. This demonstrates the limited generalization of simulated bicubic degradation. On the other hand, our proposed CDC still achieves an improvement on bicubic images and outperforms most of state-of-the-arts methods. This is also the evidence to prove the superiority and generalization of our method.
Fig.~\ref{fig:example_sr_ressult} visualizes SR results from models trained on simulated SR datasets. 
One can see that models trained on bicubic degradation produce blurry and poor SR results. This clearly demonstrates that the image degradation of the simulated SR dataset greatly hinders the performance of SR methods in real-world scenarios.

\section{Conclusion}
%
In this paper, we establish a large-scale real-world image super-resolution dataset, named DRealSR, to facilitate the further researches on realistic image degradation. To mitigate the complex and diverse image degradation, considering reconstruction difficulty of different components, we build a HGSR network with a stacked architecture and propose a Component Divide-and-Conquer model (CDC) to address the real SR problems.
CDC employs three Component-Attentive Blocks (CABs) to learn attentive masks and intermediate SR predictions with an intermediate supervision learning strategy.
Meanwhile, a Gradient-Weighted loss is proposed to drive the model to adapt learning objectives to their reconstruction difficulties.
Extensive experiments validate the challenging aspects of our DRealSR dataset related to real-world scenarios, while our divide-and-conquer solution and GW loss provide a novel impetus for the challenging real-world SR task or other low-level vision tasks.


\clearpage
%
%
\bibliographystyle{splncs04}
\bibliography{egbib}
\end{document}